# Distributional Semantics, Holism, and the Instability of Meaning

**Jumbly Grindrod**

j.grindrod@reading.ac.uk

University of Reading

**J.D. Porter**

porterjd@sas.upenn.edu

Price Lab for Digital Humanities, University of Pennsylvania

**Nat Hansen**

n.d.hansen@reading.ac.uk

University of Reading

## ABSTRACT

Large Language Models are built on the so-called *distributional semantic* approach to linguistic meaning that has the *distributional hypothesis* at its core. The distributional hypothesis involves a holistic conception of word meaning: the meaning of a word depends upon its relations to other words in the model. A standard objection to holism is the charge of instability: any change in the meaning properties of a linguistic system (a human speaker, for example) would lead to many changes or a complete change in the entire system. We examine whether the instability objection poses a problem for distributional models of meaning. First, we distinguish between distinct forms of instability that these models could exhibit, and argue that only one such form is relevant for understanding the relation between instability and communication: what we call *differential instability*. Differential instability is variation in the relative distances between points in a space, rather than variation in the absolute position of those points. We distinguish differential and absolute instability by constructing two of our own smaller language models. We demonstrate the two forms of instability by showing these models change as the corpora they are constructed from increase in size. We argue that the instability that these models display is constrained by the structure and scale of relationships between words, such that the resistance to change for a word is roughly proportional to its frequent and consistent use within the language system. The differential instability that language models exhibit allows for productive forms of meaning change while not leading to the problems raised by the instability objection.

## 1. Distributional semantic models

> *According to the first sentence of 100% of the papers I have reviewed this year, large language models have achieved amazing success in recent years. Perhaps we could settle on the abbreviation "Language Models Are Outstanding", so papers could begin "LMAO, but..." to save space.*

– Christopher Potts[1]

LMAO. In fact, LMAFO. Part of why language models have become so outstanding lies in the vector space approach to representing linguistic features. According to this approach, we represent the meaning of each linguistic unit in terms of a vector in a high-dimensional space, where those vectors are determined by the distributional properties of a large corpus. This approach to the representation of meaning has its origins in distributional semantics. This is a branch of computational linguistics which is underpinned by the *distributional hypothesis:* terms that are similar in meaning will have similar distributions across a corpus, and terms that are dissimilar in meaning will have dissimilar distributions across a corpus (Harris, 1954; Firth, 1957; Lenci, 2018; Grindrod 2023).

The simplest way to see how the distributional hypothesis can be realised is by looking at a simple "count" model. Suppose that we have a corpus of written English. We could represent the meaning of each term in that corpus in the following way. For every word, we could represent its distribution in the corpus with a list consisting of how often that word co-occurs with each expression in the corpus vocabulary (see Table 1 for a toy example of such a list). This list of numbers is a vector, with each component of the vector indicating the position of a point along a particular dimension. If our corpus vocabulary has N words in it, the vectors we produce will be in an N-dimensional space.[2]

| | "paws" | "bark" | "wing" |
|---|---|---|---|
| "dog" | 50 | 77 | 3 |
| "cat" | 48 | 4 | 2 |
| "bird" | 0 | 10 | 47 |

Table 1: Vectors in a simple count model of word distribution

[1] https://x.com/ChrisGPotts/status/1686802492104028160
[2] For an example of this kind of model constructed in Python, see: https://colab.research.google.com/drive/1ETqULGb4M5-iPA4jCSJQ9-iujeWYtuvd?usp=sharing

If the distributional hypothesis is right, expressions that are similar in meaning will have vectors that are near one another in this space. For instance, we might expect that "dog" and "cat" have more similar vectors than "bird" and "cat" do because both "cat" and "dog" appear alongside "paws", while "bird" does not. At the same time, the vectors "cat" and "dog" will be dissimilar insofar as "dog" will co-occur more often with "bark" than "cat" will.

This illustrates the distributional approach in its simplest form, but there are a range of ways in which further parameters can be introduced to improve the performance of count models. First, we left "co-occur" undefined in the last paragraph, and some decision has to be made on what co-occurrence amounts to. This is usually a case of deciding the size of the "context window" on each side of the target word (how many words on either side of the target word will count as co-occurring with it). But more sophisticated options are available, including the introduction of syntactic information or of weightings that reflect how close the co-occurrence is. It may be that rather than representing the relation between the target word and the context word in terms of bare co-occurrence, an association score such as pointwise mutual information is used, which would indicate how likely it is that one expression occurs given the occurrence of the other. There are also a number of decisions to make in pre-processing the corpus, such as whether we exclude stopwords (high-frequency expressions like "the", "a", "and", "is"), or whether we lower-case or lemmatize tokens.[3] There are also transformations that can be performed on the resultant vectors, such as a weighting function or dimensionality reduction. Note that dimensionality reduction (such as singular value decomposition) typically means that the dimensions are no longer interpretable in terms of co-occurrence with a particular expression in the corpus vocabulary. Instead, the dimensions will be tracking more abstract patterns within the distribution of words in the corpus.

The choices for these parameters are usually decided on the basis of the specific natural language processing task at hand, what the current state of the art is, and the performance of the model according to specific evaluation tasks. For example, the GLUE (General Language Understanding Evaluation) benchmark is a widely-used set of evaluation tasks that are used to test the performance of language models (Wang et al., 2018). This includes tests of whether sentences are grammatical, whether they have a positive or negative sentiment, whether pairs of sentences are similar in meaning, whether pairs of sentences have an entailment relation, and much more.[4]

Since around 2013, particularly with the introduction of Word2vec (Mikolov *et al.*, 2013), it has become common to construct vector space models of meaning via neural networks rather than the count method outlined above. On this approach, a language model is constructed by training a neural network to perform some prediction task. For instance, the Word2vec's "continuous bag of words" (CBOW) algorithm aims to predict which word will appear at a given point in a sentence, given the words that appear to the left and right of the missing word.[5] It does this in a self-supervised way, by constructing a training sample of contexts and target words and then adjusting the various weights of the neurons in the network so that the network is able to predict target words on the basis of a context to some degree of success.[6] We can understand how it does this in terms of the architecture of the neural network. The Word2vec network has three layers of neurons, an input layer, a hidden layer, and an output layer. Each neuron in both the input and output layers correspond to a different word, so if there are 10,000 different words in the corpus, then there will be 10,000 neurons at the input and output layers.[7] The hidden layer will have fewer neurons, for instance, 300. As each neuron in the hidden layer has a weighted connection to each neuron in the input layer, each word—corresponding to an input layer neuron—is captured by the weighted connections to each hidden layer neuron. In this way, each word can be represented by a vector with 300 components. Once the training phase has been completed, the connection weights have been adjusted so that the network is able, to some degree of success, to predict the missing word given the input context. This neural network approach brings with it a host of further parameters for which decisions need to be made. Two of the most important are: i) the task that the network is trained on and ii) the number of dimensions for each vector (i.e. the size of the middle layer). Again, the choices on how these parameters are set will be a matter of weighing up the state of the art, the specific aims for the model, and computational efficiency. Regarding (i) for instance, CBOW is thought to work better with more frequent terms and with larger corpora, while Skip-gram models tend to work better with smaller corpora and less frequent terms (Mikolov *et al.*, 2013).

---

[3] A lemma is the common base form of a family of expressions: [know] is the lemma for "knows", "knew", "known", "knowing", "knowed".

[4] The GLUE benchmark leaderboard can be found at: https://gluebenchmark.com/leaderboard

More recently, a great deal of the progress in NLP has been due to the rise of transformer architectures, with their characteristic use of *self-attention heads* (Vaswani et al. 2017). We will not attempt to give a complete account of how self-attention works here, but a brief indication of its motivation will be useful to see that even the recent large language models that have attracted so much attention are still distributional models at their core.

Embeddings produced by Word2vec are in an important sense completely static: a word is represented by the same vector in every sentence it appears in. But when it comes to interpretation of words in use, ordinary speakers are sensitive to many different ways in which word meaning could be affected by its position in a sentence, including, for instance, ambiguity resolution (homophony and polysemy), and anaphora resolution. The use of self-attention heads addresses this limitation. The basic idea is that each word embedding will be replaced by a new embedding that is a weighted sum of the word embeddings that represent the original sentence or text.[8] How much each word embedding contributes to the final embedding is determined by an attention score. To obtain attention scores, a matrix is produced that represents, for each word, the scaled dot product of its embedding with the other word embeddings in the sentence, which is then run through a softmax function to ensure that the attention scores are positive and sum to 1. The dot product is a similarity measure between vectors, so you might think that this just means that words closer in meaning obtain higher attention scores. However, the vectors are run through distinct linear transformations depending upon their role in the above calculation, and these essentially serve as parameters for the self-attention head to adjust how the attention scores are calculated during training.[9]

---

[5] Word2vec also has the "skip-gram" algorithm, which attempts to predict the context that a given input word appears in.

[6] Typically, the degree of success is not pre-specified. Instead, the number of training epochs that take place will be specified in advance, although some exploratory work is often done to find out how many epochs it is worth doing before the improvements diminish to a point that it is not worth the further computation.

[7] A decision may be made to restrict the network to e.g. the 10,000 most common words, so that the computational efficiency of the network is not compromised by very rare words that may not be of interest.

[8] The context window in GPT-3 is 2048 tokens long, meaning that the model is able to produce attention scores for an extended piece of text, rather than being limited to particular sentences.

Let's call this process of replacing an embedding with a weighted sum of the surrounding embeddings the *self-attention procedure*. In transformer architectures, there are multiple self-attention heads attending to different aspects of each input embedding and there are many layers of self-attention heads. That means that the self-attention procedure is repeated many times, which potentially allows each self-attention head to focus on specific features of the sentence or wider text. In theory, some might be dedicated to syntactic properties of the sentence, others might be dedicated to anaphora resolution, and others might be dedicated to ambiguity resolution (Rogers et al. 2020). The crucial point for our purposes is that while the introduction of self-attention heads allows the model to produce contextualized (rather than static) embeddings that are sensitive to the interactions between words in a sentence, this is not a fundamental departure from the core distributional model that takes as its basic starting point the idea that a word can be represented by a vector that captures its distribution across a large corpus.[10] This idea, underpinned by the distributional hypothesis, is the focus of this paper.

In the following two sections, we will outline how such distributional models are *holistic*. That means they are subject to a family of objections against holistic models of meaning, namely *instability objections*.

## 2. Meaning holism

In this paper, we will investigate the repercussions that arise from accepting the distributional view of linguistic meaning that we have seen is integral to the language models discussed in the previous section. We want to focus on the holistic nature of distributional models and consider whether these models are subject to the kind of objections that are usually raised for holistic theories of meaning. To do that, we will need to first make clear what exactly we mean by a holistic theory such that it applies to distributional models of meaning.

The history of holism about meaning in philosophy is a history of a few intertwined strands. For instance, there is a strand in philosophy of science, most closely associated with Duhem's and

---

[9] More specifically, there are three roles that the vectors play. The *Query role* is the role that the vector plays as the target word for which the attention scores are generated. The *Key role* is the role that the vector plays as a context word in generating attention scores. And the *Value role* is the role that the vector plays in being multiplied against its attention score in producing the new vector. The vector goes through distinct linear transformations as part of each role.

[10] Many of the large language models that are used in chatbots (e.g. OpenAI's GPT series) have gone through further training, often reinforcement learning through human feedback. To that extent, these models depart from the core distributional approach that is the topic of this paper.

Quine's position that observational predictions are made by an entire theory rather than by individual theoretical statements. There is the related interpretivist tradition that one finds in the work of Davidson, Lewis, and Dennett, according to which meaningful states can only be attributed as a collective (rather than individually) to their possessors according to certain principles of interpretation. There is the inferentialist tradition according to which the meanings of particular entities (whether words, sentences, concepts, or beliefs) are determined by the inferential relations that those entities hold with other meaningful entities (Harman 1974) or the meanings of speech acts are determined by inferential relations to normative attitudes of commitment and entitlement (Brandom 1994). And, more exotically (from the perspective of analytic philosophy), there is Saussure's (2013) view that the value of a word is determined by the manner in which it is contrasted with other words.[11]

In focusing on meaning holism two clarificatory points need to be made. First, because we are interested primarily in distributional models as capturing *linguistic* meaning, we will be focused on holism as a thesis that applies to linguistic entities rather than mental entities. Second, holism is sometimes taken to be a claim of priority, that the meaning of a simpler structure is derived from the meaning of a more complex structure. This is seen in Quine's claim that theoretical sentences derive their meaning from the theory as a whole, and also in the idea that words do not possess meanings independently of the sentences that they appear in.[12] However, questions of meaning priority between simpler and more complex structures is not our focus here.[13] Instead, we are interested in the idea that the meaning of a word is determined by the relations that it holds with other words. This idea is present in Saussure's (2013) notion of *value*— for instance, in his explanation of the difference in value between the French "mouton" versus the English "sheep" in terms of the fact that the English term has the contrasting term "mutton" to pick out the meat of a sheep whereas there is no corresponding term in French. The idea is also clearly present in inferentialism—in the idea that the meaning of each word is determined or constituted by the inferential relations held by sentences containing that word.

The specific form of holism we are interested in claims that the determination of a word's meaning is determined by the relations it stands in with other words. This is clearly captured in Dresner's characterisation of meaning holism:

> Relations of a certain kind (or kinds) that obtain among expressions of a given natural language (all of these expressions, or many of them) are constitutive for linguistic expressions to mean what they do. (Dresner, 2012, p. 611)

---

[11] It is in American rather than European Structuralism, and particularly in the work of Zellig Harris (1954), that we find one of the first formulations of the distributional hypothesis, although this also has to be squared against a behaviourist theme in his work that meaning (in at least some senses of the term) is not the proper subject of linguistics.

[12] This of course comes very close to Frege's context principle: "The meaning of a word must be asked for in the context of a proposition, not in isolation" (Frege, 1997, p. 90). However, the extent to which Frege was thereby committed to some kind of holism is a matter of debate (Janssen, 2001; Pelletier, 2001).

[13] While holistic views frequently are committed to the claim that word meaning is derived from sentence meaning, we will see that the type of holism suggested by distributional models do not fall within this category. Distributional models in the first instance provide a way of representing word meaning, with sentence meaning left unaccounted for without some further procedure. This is true even of transformer models, which output contextualized embeddings for each word in a text. Thanks to Kamil Lemanek for emphasising this point.

For our purposes, the distributional models outlined in the previous section fall under this definition, where the relations in question will be distributional: that is, how expressions are distributed across a corpus in terms of the other expressions that they frequently co-occur with. Representing the meaning of a word in terms of its distribution is to represent it as part of a holistic system (Grindrod 2023).

Of course, one may be skeptical that distributional models do properly represent the meanings of expressions. It may be that distributional models merely provide a way of simulating certain meaning-related phenomena, while genuine meaning properties are determined in ways that make no appeal to distributions (Bender and Koller 2020). Similarly, it might be thought that the determination relation actually runs in the opposite direction—that the meanings of particular expressions determines how the expression is used and thus its distribution across a corpus. On this picture, distributional models would constitute a form of reverse engineering (Lenci, 2008, p. 14; Boleda, 2020, p. 2). We acknowledge that any complete defence of a distributional theory of meaning would need to address these possibilities.[14] But we will put these concerns to one side because we're interested in using these holistic models to illuminate some classic philosophical worries about holistic conceptions of meaning.

## 3. The instability objection to meaning holism

Meaning holism has faced several criticisms. For example, Fodor and Lepore (1992, ch. 6) have argued that it is inconsistent with facts about compositionality, and Dummett (1993) has argued that it cannot provide a plausible account of language learning. Arguably the most pressing and well-known objection is the *instability objection* (Lormand 1996). The problem in its most basic form is simple: if the meaning of a word is determined by its relations with other words, then a relevant change to one word will lead to meaning changes in many other words. The ensuing instability of word meaning is incompatible with the ease of successful communication and language acquisition that human languages possess. In its strongest form, the instability objection claims that any change in the language would lead to a complete change in the language. Pagin (2008) labels this the "total change thesis". For the total change thesis to be the case, it would need to be that the language is, as it were, *completely holistic* insofar as the meaning of any word is determined by its relations to all other words (or at least that there would be a path of determination from any word to any other word). But even if a language system falls short of being completely holistic, the instability that would come with a highly interconnected system may still spell trouble for any account of linguistic meaning that requires word meaning to be stable.

Consider the problems that instability poses for communication. A plausible necessary condition for successful communication is that speaker and hearer identify the same meaning when a particular word or sentence is used. But if the meaning of a word is determined by a complex network of relations to all other words in a linguistic system, then it seems as though speaker and hearer will only be able to alight upon the same meaning if the linguistic systems that they employ are completely identical. But given differences between individuals, this is implausible. Not only do our understandings of particular words naturally vary (whether through error or otherwise), but the size of our vocabularies vary as well. This fact alone would seem to prevent the possibility of a hearer and speaker extracting the same meaning from an uttered sentence if word meaning is holistic. So, if the holistic picture is right, it would seem doubtful that anyone ever successfully communicates.[15] Fodor and Lepore take the problem of instability to generalize away from the specific issue of communication. For instance, they argue that the instability that holism brings makes generalizations about intentional mental states impossible. However, such objections only apply to distributional models if they were to be used in that specific theoretical role. We acknowledge that the problem of instability may lead to problems for a theory of meaning beyond communication, but we will use the issue of communication to frame our discussion in what follows.

[14] For a detailed defence of the use of language models to inform theories of meaning, see: Grindrod (2024, forthcoming).

A natural response to the instability objection is to appeal to the notion of *meaning similarity*. The thought would be that the problem instability poses for communication only has bite if we require meaning identity rather than meaning similarity for successful communication. If this requirement is loosened so that communication is successful when speaker and hearer share relevantly similar meanings, then communicators with two different linguistic systems can still successfully communicate if the meanings they attach to expressions are similar enough. We have already seen, in §1, that meaning similarity is a central part of distributional models: such models give us a precise way of measuring the similarity between the meaning of expressions, in terms of their proximity in vector space.

Using the resources of distributional semantics, in the following sections we will develop a similarity-based response to the instability objection in two parts. First, in order to appeal to meaning similarity, we need to specify the precise sense in which two expressions can be similar to one another. Once we have a clear account of meaning similarity in place, we will then show, using two models, the extent to which distributional models are unstable. We will argue that the instability that they display is not problematic. This is partly because the instability reflects subtle changes in word meaning, but also because the instability is tempered as the models train on larger and larger sets of data.

## 4. Meaning similarity

The idea of meaning similarity has come under sustained philosophical criticism from Fodor and Lepore (Fodor and Lepore, 1992, 1996, 1999). Their proximal target is Paul Churchland's (1979) vector-based semantics for neural processes. Churchland details how groups of neurons process visual information via a neural network that will be sensitive to features of visual information. He represents the activation patterns of neurons in the group using a vector space. A discussion of the details of Churchland's view will take us too far afield from the question of distributional models, but we can extract the objections that Fodor and Lepore made against his vector-space view of meaning similarity for the purposes of applying it to distributional models of word meaning.

[15] We will be using a simplified picture of communicative success where the question is just whether speaker and hearer identify the same or similar content. This leaves out other important questions such as the role of speech acts in communicative success. It may be thought that in adopting this view of communication, we are already begging the question against the holist to some extent. Brandom (2007, p. 666-670) for instance, argues that the inferentialist can adopt a view of communication according to which it amounts to speaker and hearer successfully navigating the commitments and entitlements that each are subject to, and where this can't be understood as each alighting upon the same or similar content. Fodor and Lepore (2007, p. 684-688) argue that such views still depend upon and lack any clear account of meaning similarity. This paper explores a way of developing a clear account of meaning similarity using distributional models, and it can do so without adopting the inferentialist-friendly account of communication. Thanks to Kamil Lemanek for raising this point.

One reason to be skeptical of the theoretical usefulness of meaning similarity is that some similarity relation (i.e. similarity in *some* respect) will always hold between any two objects, and this applies to meaning as much as to anything else (Goodman, 1972; Fodor and Lepore, 1999, p. 392 ff.). If we are to make use of a similarity relation in our account of successful communication, we need to state in what respect two meanings must be similar or else we would end up with a vacuous view that holds that every communicative exchange is successful because every meaning is similar to every other in some respect.

Vector space semantics provides the resources to respond to this objection. If meaning similarity is just relative proximity in a vector space, then it can be represented mathematically as the distance between points in an N-dimensonal space (in terms of cosine similarity or Euclidean distance, for example). However, Fodor and Lepore (1992, ch. 7) argue that the problem of specifying a relevant measure of similarity re-occurs as the question of how the dimensions for a vector space are selected. The claim that two words are located in similar parts of a semantic space first depends on the semantic space being defined in terms of a set of dimensions. In making this point, Fodor and Lepore envision a semantic space (akin to Churchland's) where the dimensions are semantically interpretable; for example one dimension might stand for blueness and another might stand for height, and so on. Because Fodor and Lepore envision the semantic space in this way they then phrase their objection as the claim that the vector space theorist must draw a principled distinction between those semantic properties that serve as dimensions of the space, and those semantic properties that are represented in the semantic space.[16] But in the distributional models we are considering, the dimensions in question are not semantically evaluable. Even for simple count models, prior to any dimensionality reduction, the dimensions would stand for co-occurrence statistics with other words, which is a distributional property of a term, not a semantic property. And on any model that has been through dimensionality reduction or that is produced using a neural network, the dimensions are not even interpretable in those terms.[17]

So the distributional theorist does not face the challenge of drawing a distinction between those semantic properties that serve as dimensions and those that do not. But Fodor & Lepore's challenge raises a more general question. Given that meaning similarity is defined in terms of a space, and the space itself is defined in terms of its dimensionality, how do we decide upon the dimensionality of the space? It may be that in a space with dimensionality N, a particular pair of words are very close together, but that in a distinct space of dimensionality N+1 they end up quite far apart. In a case like this, do we treat the pair of words as similar or dissimilar in meaning?

---

[16] They view this as essentially a new version of the older problem of how to draw a principled analytic/synthetic distinction. The idea is that we would still need to identify a set of basic semantic properties from which the non-basic semantic properties are composed (where analytic statements would capture those compositions).

[17] For models that have been through singular value decomposition (a common form of dimensionality reduction), the dimensions will capture the dominant degrees of variation in the proximity between words, as represented in the original non-reduced space. For smaller neural network models (e.g. Word2vec), the dimensions stand for the activation patterns of neurons in the hidden layer. For large language transformer models, it is difficult to say much at all about what the dimensions stand for.

This is not so much an objection against the distributional approach as much as it is a specification of a parameter. It is already widely recognized that the number of dimensions in a semantic space will affect the final model, and it is generally thought that a higher-dimensional model will lead to more accurate results, albeit at the cost of computational efficiency (Mikolov *et al.*, 2013, 7). But the dimensionality of the space may end up being unnecessarily large. You can plot a circle in a 3-dimensional space, but any number of dimensions greater than 2 will have dimensions that are redundant for that task. Similarly, the way that words are plotted within the language model will form a cloud with a particular shape, and there will be a certain number of dimensions required to accurately capture it. If the dimensionality is too low, then certain features of the word cloud will be left out (a limitation that is accepted when 3D or 2D visualisations of semantic spaces are constructed). If dimensionality is too high, then the space is redundantly large. But this is something that can be explored empirically, by exploring the effect of dimensionality on the performance of the model in evaluation (for example in the GLUE benchmark evaluation mentioned earlier).

A related objection is that this approach doesn't allow for similarity measurements across semantic spaces that differ in dimensionality (Fodor and Lepore, 1999, p. 394). Similarity measures within a space obviously say nothing about how vectors from different spaces could be compared.[18] This brings us to an important point about how meaning is typically viewed as represented within a vector space.

To get beyond the problem of using different spaces with potentially different dimensions as representations of meaning similarity, meaning similarity is treated *differentially*. On this approach, the meaning of a word is not understood primarily in terms of the specific coordinates of the vector but in terms of other nearby words.[19] These relations between words could be represented in more than one way within a single space and could be represented across spaces of different dimensionalities. For instance, *any* rigid rotation of the full set of vectors about the origin of the space (i.e. the zero point for all dimensions) will preserve all the similarity measures we have mentioned here (i.e. cosine, Euclidean, city-block).

[18] Churchland was sensitive to this concern as it applied to his vector-space account of neural activity. Given that we can expect a fair amount of neural diversity across individuals, the worry would be that vector space models do not provide a way of making generalizations about neural activity and could only provide a way of capturing the neural activity of a single individual. In Churchland (1998) he proposes a mathematical model for capturing similarity across spaces of different dimensionalities.

[19] This differential approach to meaning in semantic spaces can provide surprising findings. The most well-discussed are the so-called *analogy findings*, initially presented in Mikolov, Yih and Zweig (2013). Roughly, this is the idea that if you subtract the vector for "man" from the vector for "king", and then add the vector for "woman", you get a vector very close to the vector for "queen". This suggests that gender has been encoded into the differential structure of the space, and Mikolov, Yih, and Zweig explore how something similar holds for many different semantic and syntactic regularities. Using a word2vec model built on 20th century novels, we found analogies that seemed to capture properties of concepts (shirt - arms + legs = trousers), national differences (coffee - gray + grey = tea ), and some more surprising similarities (guitar - briefcase + desk = piano).

A standard operation in exploring the vector of a particular word is to consider its nearest neighbours. The useful point about this differential approach is that understanding the meaning of a word in terms of its nearest neighbours is something that can be done across spaces of different dimensionalities. So if a word keeps its 100 nearest neighbours across two different semantic spaces, this is reason to think that the word is represented similarly in those two spaces. This form of reasoning runs in the other direction when distributional semantics is used to investigate meaning change over time. In Hamilton *et al.'s* (2016) work on statistical laws of semantic change, they exploit this idea in the following way: they sought to produce a semantic space that maps how expressions have changed in meaning over time. They did this by first producing several semantic spaces trained on corpora from different periods in history. They then needed a way to align the spaces so that the dimensions across each can be identified. They did this using a formal method for mapping one matrix onto another known as *orthogonal Procrustes*. Put in spatial terms, this is a way of mapping one vector space onto another such that the two sets of vectors align with one another as closely as possible. They then tested whether this method had been successful by checking whether the space correctly captured known semantic changes in terms of the nearest neighbour lists across time. For instance, it is known that the meaning of "awful" has shifted from meaning roughly *impressive,* or *awe-inspiring* to its current meaning of *terrible* or *bad.* So Hamilton *et al.* evaluated whether they had successfully aligned the numerous semantic spaces by seeing whether the nearest neighbour list of "awful" reflects this known shift. In that respect, they relied upon this differential view of meaning to check that the spaces had been aligned properly.

Understanding the meanings of terms differentially helps make clear that the "space" described in a distributional model is metaphorical, arising only as a way to describe the relationships among the words. For instance, in a simple "count" model like the one depicted in figure 1, the dimensions are just the word types, so there is nothing external to word usage that situates any of the terms. In this sense, Mary Hesse's (1974; 1988) "network model" of holism may be a more apt metaphor than "space" for thinking about distributional models. Since the data consists solely of the relationships of each word to each other word, it can be depicted, with no loss of information, as a network graph, where each word type is a node, and edges connect two nodes if the corresponding words are used near each other in the corpus. These edges can be weighted to reflect the number of co-occurrences observed. In a network graph, space doesn't mean much. It doesn't matter that a node is "here" or "there"; instead, it matters how the node connects to other nodes, which nodes are along the shortest routes from one node to another, and so on. The network still enables rigorous quantitative analysis (e.g. the detection of communities, measures of robustness, ranking nodes by total edge weights, and so on), but it departs from the notion that a node is situated against a pre-existing background "space"—there is no basic dimension like Fodor's and Lepore's "blueness" here, because there are no dimensions at all.

It is not clear that Hesse intends her model quite so literally, and often she is describing a network not of words but of beliefs, along the lines of Quine's "web of belief".[20] In fact, networks are closely associated with holism in general. As Fodor and Lepore write:

> The metaphor that often goes with semantic holism is as follows: A theory is a sort of network, in which the statements are the nodes and the semantically salient relations among the statements are the paths. The meaning of a statement is its position in the network…" (1992, 42)

So while Fodor and Lepore's point about dimensionality selection is relevant, treating meaning differentially—where meaning is understood primarily in terms of the relations between words—provides a way of thinking about meaning similarity that is not tied to a specific set of dimensions. Nevertheless, there are still issues of meaning instability that need to be confronted. It may be that language models prove to exhibit a high degree of variation in the similarity relations between words as the corpus data they are trained on changes. If that is the case, then it seems that some form of the instability objection will apply to distributional semantic models.

## 4. Modelling the instability objection

In this section, we will explore the extent to which language models do exhibit instability in their representations of word meanings, even once we have acknowledged that meaning similarity is to be understood differentially in terms of word-to-word relations. We will be focused on the extent to which expansion of the corpus affects the way in which meaning is represented. We will first explore this by constructing a toy model that uses two novels as its corpus, before looking at a more sophisticated model built using Word2vec on a much larger corpus consisting of Wikipedia and the *Stanford Encyclopedia of Philosophy*. By building some actual models, we can make the evaluation of instability more concrete.

We will first run through the basic steps of constructing a count model on a small corpus. Our initial corpus for this example will be a novel. Of course, novels are in many respects quite unlike the type of ordinary language use that a model that aims to capture linguistic meaning would be built from. Novels are typically long, fictional, self-consciously stylized, and subject to extensive revision by authors and editors, a combination which sets them apart from corpora like transcribed speech, newspaper articles, or social media (or corpora that combine these and other forms). However, for the purposes of illustrating how expanding a corpus with a particular type of text may affect the meanings of particular expressions within the model, novels can be particularly useful, because they are texts for which we have a great deal of supplementary information (i.e. we can know about their subject matter, style, age, etc.), which can help us focus our attention away from these potential confounding variables and toward the quantitative results afforded by the model.

Our initial corpus for this example is Gertrude Stein's novel *The Making of Americans*. Two features make this text well-suited for the task: It is very long, and it has a low type-token ratio.

---

[20] Quine, moreover, seems to have been ambivalent about the network metaphor, often using metaphors like "fabric" or "field of force" to describe his holist views in "Two Dogmas of Belief", and elsewhere attributing to his co-author J.S. Ullian the titular metaphor of *The Web of Belief* (Quine, 1991).

*The Making of Americans* is about 522,000 words long, but contains only 5,317 unique words, for a type-token ratio of 5780 / 522,000, or about .01.[21] This means we get a lot of examples of each word in action, but the number of word-to-word relations is relatively low. The number of word-to-word relations in particular scales up quickly as the vocabulary gets more diverse, since it reflects the square of the total vocabulary (imagine a grid showing the relationship of each word to each other word—each new word adds an entire row and column).

The basic data we gather from *The Making of Americans* is this: For each target word T, we look ten words to the left and ten to the right; for each collocate C that appears in that window, we increment the number attached to C for T.[22] So, if we want to see the words that most frequently co-occur with, say, "know", we get the following:

| **Collocate** | **Co-occurences** |
|---|---:|
| i | 1201 |
| to | 1095 |
| it | 1059 |
| of | 968 |
| one | 966 |
| and | 956 |
| in | 933 |
| them | 808 |
| that | 629 |
| the | 622 |

Table 2: Most frequent collocates with "know" in *The Making of Americans* corpus

These results are not very informative in this state, since they just show that common words occur frequently ("to" and "it" don't necessarily have much to do with "know" specifically, they appear frequently near everything). This is where we can turn to a distributional model. We

[21] Stein's famously repetitive style is a significant historical outlier in this respect; compare *Bleak House* (324,000 words, about 15,000 of them unique), *Moby-Dick* (218,000 and 17,000), or *Middlemarch* (324,000 and 15,000). In fact, even *The Great Gatsby* has more unique words than *The Making of Americans*, in spite of being less than 10% as long (50,000 words long and 5,780 unique words).

[22] We leave the token T out of the window, though the same type may occur elsewhere within the window, in which case we would augment the relationship between T and its own type (the word "rose" is a collocate of the word "rose" three times in "Rose is a rose is a rose is a rose").

create an array where each target word T gets a row. The columns capture the number of co-occurrences between T and every collocate C, as follows:

| | i | to | it | of | one | and | in | them | that | … |
|---|---|---|---|---|---|---|---|---|---|---|
| know | 1201 | 1095 | 1059 | 968 | 966 | 956 | 933 | 808 | 629 | … |

Table 3: Most frequent collocates with "know" in *The Making of Americans* corpus, target words in rows

We can then treat these columns as dimensions and calculate the distance between any two words from the rows, just as we would if the rows contained cities and the columns were latitude and longitude. Here are the closest terms to 'know' in terms of cosine similarity:[23]

| **Term** | **Cosine Similarity** |
|---|---|
| know | 1 |
| understand | 0.95231 |
| it | 0.94286 |
| see | 0.93759 |
| think | 0.93293 |
| feel | 0.92994 |
| like | 0.91969 |
| i | 0.91810 |
| love | 0.91631 |
| say | 0.91139 |

Table 4: Closest terms to "know" measured by cosine similarity in *The Making of Americans*

Here we see several other verbs of cognition and perception, in general a much clearer picture of terms that share conceptual space with, or are used similarly with, "know". Even in this relatively small corpus, consisting of just one novel, the distributional semantic approach captures something meaningful about how words are used.

Given the differential view of meaning that is typical across distributional approaches, it is nearest neighbour lists such as these that reflect the meanings of particular expressions, rather than the specific point in space that the expression occupies. What happens to nearest neighbour lists if we add some additional language to the model? To check, we ran the same calculations on a corpus that consists of *The Making of Americans* and Ernest Hemingway's short story "A Clean, Well-Lighted Place". This is a useful supplementary text for two reasons: It is very short (just under 1500 words) and the type-token ratio is, again, very low (just 412 unique words).[24] Both features suggest that adding this story to the corpus should only slightly affect the relationships between words established in the much longer novel, which gives us an initial test of the instability worry that changing any of the relations in a linguistic system changes the whole thing.

[23] The pdist module we use in Python returns cosine distance. Here we report cosine similarity, which is just 1 - cosine distance, i.e. a higher number means greater similarity. This makes the charts in our toy model a little easier to compare with the word2vec similarity scores.

With "A Clean, Well-Lighted Place" in the mix, we see a few different kinds of change to the meanings of terms in our model. Some words appear for the first time: *The Making of Americans* does not include the word "nada", but Hemingway uses it 21 times in his short story. This new term still has a relationship to words used only in Stein's novel, e.g. it has a cosine similarity of .17188 to "hersland", the last name of a family in *The Making of Americans*.[25] That's because many words are shared between the two texts, and those shared terms function as connective tissue for unique terms. Two other kinds of changes, however, get more to the heart of the instability worries considered above.

First, there are terms that do change a lot when Hemingway's text is added. Hemingway uses the word "glass" 5 times in his short story, compared with its 2 uses in *The Making of Americans*. That's a substantial percentage of all uses in the total corpus, so we might expect to see a shift in the meaning of "glass" in the version of the corpus that includes the short story. And we do:

| **Term** | **Cosine similarity** | | **Term** | **Cosine similarity** |
|---|---|---|---|---|
| glass | 1 | | glass | 1 |
| chandeliers | 0.72183 | | saucer | 0.70040 |
| prisms | 0.66340 | | waiter | 0.69522 |
| brocade | 0.66189 | | table | 0.67740 |
| couches | 0.62936 | | leaves | 0.67237 |
| paintings | 0.61804 | | sun | 0.66906 |
| covered | 0.59258 | | wind | 0.66725 |
| blue | 0.57989 | | rain | 0.65831 |
| peasants | 0.56695 | | ground | 0.65248 |

[24] The latter stems partly from Stein's immense influence on Hemingway's style; early in his career, he spent a lot of time with her and learned from her approach to writing. This connection between his work and hers is another reason we might expect adding this short story to the corpus to operate like the addition of a new discussion in an ongoing conversation—more language in a context where the speakers expect to be on the same page.

[25] All words are rendered in lowercase for the purposes of making the model; the name is rendered as "Hersland" in the text of the novel.

| grace | 0.56356 | | autumn | 0.64633 |
|---|---|---|---|---|

Table 5: Most similar words for “glass” in the original corpus (left) and the augmented corpus (right)

In this case, the change is substantial and easy to interpret: “A Clean, Well-Lighted Place” is set at a café, so the term “glass” is used in the sense of “a thing to drink alcohol from”, and the fact that the term is so much more significant in the short story produces a set of top terms and distances, that is quite distinct from “glass” in the original *Making of Americans* corpus.[26] On this basis, it seems fair to say that “glass” *means* something different in the augmented corpus than it did when it contained only the text of Stein’s novel. Clearly this is partly due to the ambiguity of the term that this model is blind to (i.e. between the adjective, the mass noun, and the count noun—with Hemingway primarily using the count noun form). Ambiguity of this kind can be captured in more sophisticated models (e.g. by using parts-of-speech tagged corpora or by using contextualized embeddings). Nevertheless, the point remains that the different way in which the term is used in “A Clean, Well-Lighted Place” has had a significant effect on the way that the expression’s meaning is represented. In effect, expanding the corpus with the Hemingway text is like moving from a language where “glass” is rarely if ever used to pick out drinking receptacles to a language where this is the most common form of use, and the model captures that.

More common, however, is a second and subtler kind of change. Consider “know” again: it is an extremely common word in Stein’s text (1,405 occurrences), while it is only used by Hemingway 3 times. Hemingway’s short story therefore does not contribute much to the overall stock of “know” uses, but then again that probably mimics standard types of modest additions to linguistic systems better than the “glass” example did—most new utterances make relatively small changes to the total linguistic picture. And we can observe the relatively small effect of adding Hemingway’s text to the corpus. Here are the two tables comparing the way the meaning of “know” is represented before and after the addition:

| **Term** | **Cosine similarity** | | **Term** | **Cosine similarity** |
|---|---|---|---|---|
| know | 1 | | know | 1 |
| understand | 0.95231 | | understand | 0.95284 |
| it | 0.94286 | | it | 0.94282 |
| see | 0.93759 | | see | 0.93767 |
| think | 0.93293 | | think | 0.93301 |
| feel | 0.92994 | | feel | 0.92995 |

[26] Often the changes are less interpretable, since words are entangled with each other in complex ways.

| like | 0.91969 | | like | 0.91988 |
|---|---|---|---|---|
| I | 0.91810 | | i | 0.91824 |
| love | 0.91631 | | love | 0.91621 |
| say | 0.91139 | | say | 0.91149 |

Table 6: Most similar words for "know" in the original corpus (left) and the augmented corpus (right)

In one sense, this table captures exactly how the instability charge could be levelled against the distributional approach. All the values in the cosine similarity columns have changed—even Hemingway's miniscule contribution to the overall corpus has altered them. At the same time, these changes are small, rounding errors compared to what we saw with "glass". More important, however, is the fact that the list of top terms is the same in both tables: the same words appear in the same order. If meaning is understood in terms of absolute location in vector space, the meaning of "know" has changed across the two tables; its cosine similarity to "understand" is now 0.00054 closer, among other things. But if we instead understand meaning according to the differential picture outlined in the previous section, these two tables depict near-identical meanings. One way of understanding this is that the addition of new material has meant that the precise point in space that "know" occupies has changed, but it remains in the same region. But, more importantly, given that meaning on the distributional approach is best understood not as tied to a specific point in space, but in terms of word-to-word relations, the meaning of "know" remains stable even after the change in the corpus, as captured by the fact that its relations with its nearest neighbours remains stable.

We can start to see how this differential view of meaning is well-suited for capturing several intuitive features of language and its capacity to change. Intense usage of a new term (like "nada" in our toy corpus) or of a previously rarely used term (like "glass") may create big local changes while leaving most of the system in roughly the same order as before: changes to "glass" had little effect on "know", for example. A few new uses of a very common term may do little to change the meaning of that term *or* the system as a whole, although with enough time and new use even well-entrenched words might adjust. For example, consider the changes in meaning for very common words like 'wanting' and 'starting' over the course of the 20th century, which used to mean primarily 'lacking' and 'snapping to attention' and now typically mean 'desiring' and 'beginning' (Hamilton et al, 2016, p. 1494). Finally, in a communicative exchange, two speakers may arrive at a conversation with somewhat different understandings of word meanings. "Know" will carry some different idiosyncratic associations for one speaker than for another, and a conversation between them may modestly tweak "know" for each interlocutor. In spite of those changes, we may never feel the meaning of the word, much less the language as a whole, to be radically unstable, as the meanings of all other expressions may remain fixed. To put this in network terms again, Hesse writes, "there is no *a priori* guarantee that two persons brought up in the same language community will use their words with the same meanings in all situations…. but … rational communication can take place in intersections" (1974, 37). In graph-theoretical terms, the intersection of my network and yours consists of all the nodes and edges we share. Necessarily, then, the intersection of our personal language models after any communicative utterance will include all the words in that utterance (since we have both heard them) along with all of the connections of the words in that utterance. It will also contain any words we have both ever heard, and the shared connection patterns among those words. We may each have somewhat different edge weights and some unique nodes in our personal models, but the nearest neighbors (and shortest paths) for the words in the intersection graph (i.e. a graph of all the connections that our two networks both possess) can still be quite similar, or even exactly the same as those in our personal models (perhaps differing in specific edge weights, akin to the small variations in cosine similarity above, but still consisting of the same word lists and structures of connection). To the extent that this holds, we will understand the meaning of the words in the utterance the same way. The notion that meaning depends less on exact isomorphisms than on proximity to a location, defined by the relationship of all to all in the system of words, allows for a stable-enough account of meaning that accommodates distinctions between speakers or discourses and suggests empirically feasible avenues for change (or the lack thereof) over time.[27]

In a larger language model our simple examples become more difficult to interpret.[28] But a simple examination suggests that the principle essentially holds at larger scales. We built a Word2vec CBOW model on a random sample of Wikipedia articles, consisting of 372,194 articles and about 143,000,000 words. We then built a second model consisting of the same articles plus the entire Stanford Encyclopedia of Philosophy, (SEP) 1,763 articles and about 21,000,000 words. In adding the SEP to our original corpus, we expect there to be significant changes in the meaning of expressions as we saw in our toy example, especially with respect to philosophical terms of art. That said, this is not an apples-to-apples comparison with the toy example. First, in percentage terms, the SEP makes up a much larger portion of the combined corpus than "A Clean, Well-Lighted Place" did when we added it to *The Making of Americans* (about a 15% increase in word count for SEP, vs just 0.3% for the Hemingway).[29] Second, it is important to acknowledge that parts of the Word2vec process are random, and so running the same algorithm on the same corpus multiple times will produce different results each time.[30] At the level of the components of the individual vectors, it is practically guaranteed that there will be variation across multiple iterations of the algorithm on the same corpus. But the important question for our purpose is whether this also leads to variation in the nearest neighbour lists for particular expressions.

---

[27] The limitations of collocate-based models of distributional semantics complicate these ideas (for instance, our method here totally ignores polysemous terms), but the basic principle does not depend on the simplicity of this example. Even in a very complex holistic system of meaning, two utterances of "know" would not have to be identical (in whatever terms used to measure their meaning) for their meaning to be mutually intelligible.

[28] Here we mean "larger" with respect to the corpus used to build the model, but it is important to note that people sometimes discuss model size in terms of other metrics, especially the number of parameters used in training the model.

[29] In terms of new types, the SEP adds 34,330 new words to the 1,202,353 in the Wikipedia corpus, an increase of 2.9%. The Hemingway story adds 94 new words to Stein's 5,317, an increase of 1.8%.

[30] Specifically, the initial weightings for the connections between neurons is assigned randomly. For a word of caution about the consistency of results in word vector models, see Antoniak and Mimno (2018).

All that said, the effects of adding the SEP to the Wikipedia corpus are roughly in line with what we saw in the toy example. For a word like "atomism" that appears rarely in the Wikipedia articles (9 times) and relatively frequently in the SEP (662 times), the top terms and their weights change noticeably:

| **Term** | **Similarity** | | **Term** | **Similarity** |
|---|---|---|---|---|
| hermeneutics | 0.62644 | | Hermeneutics | 0.64373 |
| marxian | 0.62019 | | literalism | 0.60139 |
| literalism | 0.60581 | | dialectical | 0.59741 |
| teleological | 0.59551 | | aristotelian | 0.59328 |
| structuralist | 0.59108 | | common-sense | 0.59327 |
| hegelian | 0.58919 | | dialectic | 0.59256 |
| dialectics | 0.58752 | | marxian | 0.59049 |
| dialectical | 0.58671 | | stoics | 0.59037 |
| aristotelian | 0.58567 | | metaphysics | 0.59004 |
| dialectic | 0.58515 | | teleological | 0.58983 |

Table 7: Top terms for "atomism" in the Wikipedia corpus (left) and that corpus augmented with the SEP (right)

But for a common word like "can", the differences are minute, even though the SEP texts grew the corpus by 15%:

| **Term** | **Similarity** | | **Term** | **Similarity** |
|---|---|---|---|---|
| could | 0.90703 | | could | 0.90923 |
| must | 0.89361 | | must | 0.89654 |
| will | 0.87542 | | will | 0.87336 |
| should | 0.86157 | | should | 0.86216 |
| might | 0.84092 | | might | 0.84254 |

| would | 0.77246 | | would | 0.77117 |
|---|---|---|---|---|
| shall | 0.74333 | | shall | 0.74569 |
| tends | 0.71550 | | tends | 0.71480 |
| tend | 0.65775 | | allows | 0.65867 |
| allows | 0.65653 | | tend | 0.65672 |

Table 8: Top terms for "can" in the Wikipedia corpus (left) and that corpus augmented with the SEP (right)

Though these results must be taken with a grain of salt, they tend toward the conclusions we arrived at with the toy model. New discourse—even quite a lot of it, and even when it is contextually quite distinct from "existing" discourse—may or may not change the meaning of words, so long as we understand that meaning to derive not from some ever-changing pinpoint location in space, but from a relative relationship to terms taken as a whole. Changing discourse *might* change the meaning of a word like "glass" or "atomism"—and surely we want our models of linguistic meaning to have that capacity. But it might leave "know" or "can" relatively unchanged, as it does here. Moreover, as the model incorporates more examples of language, the difficulty of finding cases like our "glass" example increases drastically. Antoniak and Mimno (2018) observe that larger corpora tend to correlate with more stable language models across a variety of algorithmic approaches, and this makes intuitive sense if we think about what instability means for these models. This may be clearest in the network metaphor. If the position of some node P consists of its relationship to all other nodes, then early on, as observations about connections come pouring in, P's location may fluctuate wildly—because *all* the nodes are moving around. As those other nodes lock into place (e.g., we get many observations about "know" or "can"), it gets more and more difficult for P to change relative to everything, even if P is a seldom-used word. Some nodes are effectively knotted into regions, and P is now limited to moving relative to those stable structures. The Wikipedia corpus may not contain much information about "atomism", but it does contain information about many of the words atomism is near, and the words those neighbors are near, and so on. As the corpus approaches the totality of linguistic experience, it becomes quite resistant to big changes from small sources. At the scale of actual discourse—finite words, ranked relationships to other words—meaning is stable. Or, at least, stable enough.

## 5. The sufficiency objection

We take these cases to show that the instability that one finds in distributional models does not have to be fatal to communication. Even when we significantly expand the corpus that the model uses (as in our second model), we find that some expressions will preserve their meaning while others will alter their meaning in ways that are unsurprising. In fact, we are inclined to adopt the stronger position that the instability is in fact not a drawback to the approach but is a positive feature. It allows for the detection of very subtle shifts in meaning that would otherwise be difficult to spot. This approach may well prove very useful to e.g. the question of communicative success insofar as it will capture differences in how a speaker and hearer understand the meaning of a term. Or to use a different example, this approach may have real utility in mapping the way that the meanings of particular words change over time, as the work of (Hamilton et. al. 2016) and others has already shown. It is precisely the instability (or sensitivity) in the approach that allows for this.

But there is still a further objection that could be levelled against the distributional approach. Our response to the instability objection has been first to clarify the notion of similarity that is being employed in distributional models. It is crucial to note that meaning similarity is understood differentially, as determined by word-to-word relations, rather than by a specific point in the vector space. Focusing on differential stability disarms the objections that Fodor and Lepore levelled against vector space semantics. We have then sought to show that models become more stable as they increase in size and further that the instability that remains is an important source of information. But at this stage, a critic may protest that we still need some account of *how similar* meanings need to be between what the speaker says and what the hearer receives to explain communicative success. Without such an account, there is no way of differentiating between successful and unsuccessful communicative exchanges.

There are two possible responses to this objection. First, the idea that there needs to be some threshold between successful and unsuccessful communication that comes straightforwardly from this account of meaning similarity may be mistaken. An alternative position is that communicative success is itself a matter of degree. In every communicative exchange there is a discrepancy in the meanings attached to expressions by speaker and hearer, even if it's very small. Most of the time this goes unnoticed. Sometimes, however, the discrepancy leads to real world differences. You ask if I like peas as you are cooking dinner, and I say yes. When dinner comes, I see that you have used marrowfat peas, which I don't really like. The discrepancy in how we understood "peas" has had noticeable consequences, and I rue the communicative failure as I dodge the peas on my plate. So communicative failure is only labelled as such when it is noticed, and of course it is less likely to be noticed as meaning similarity between the signal transmitted and received increases.[31]

An alternative response would be to just treat the question of the level of similarity sufficient for communicative success as an empirical question. It may be that there is in fact a level of dissimilarity that we are not willing to tolerate as communicators, and maybe in some cases even very small differences in meaning would lead to failures of communication. But this would not be something that we could tell from the structure of the model. Instead, this is a question that would need to be answered by looking at communicators' reactions to communicative exchanges and noting where they complain about communicative breakdown. This response is modelled after Haugeland's (1979) account of what it would take to "empirically defend the claim that a given (strange) object plays chess", which crucially involves "let[ting] some skeptics play chess with it". According to Haugeland, this requirement "bears all the empirical weight, for satisfying it amounts to public *observation* that the object really does play chess" (p. 620), or in this case, that two linguistic systems (people, LLMs), share meanings that are similar enough for successful communication.

[31] This example involves a discrepancy that involves a difference in extension, but it need not. The same point could be made using a discrepancy in connotation, for example.

## 6. Conclusion

In this paper, we have outlined the distributional approach to meaning that one finds in recent vector space models. We have shown how these models represent a form of meaning holism, and as such they are initially subject to the instability objection (or family of objections) that philosophers have forcefully pressed against different types of holist theories of meaning. We have argued however, that the distributional semanticist can respond to these objections. The core of this response lies in a proper appreciation of the differential way in which meaning is treated in these models, coupled with an illustration of the kind of instability that these models are subject to. What we find is that the instability can be a virtue.

The view of meaning we have defended here is not without precedent in the philosophical literature. For instance, we have already seen that Hesse defended a network theory of meaning, according to which the "use of a predicate in a new situation in principle shifts, however little, the meaning of every other word and sentence in the language" (1988, 5). Rorty, in a comparison of Hesse's network view with Davidson's truth conditional theory of meaning used an analogy with physical theories of gravity to capture a difference between the two positions:

> Davidson's resistance to this 'network' view can be put in terms of an analogy with dynamics. In the case of gravitational effects of the movements of very small and faraway particles, physicists must simply disregard insensible perturbations and concentrate on relatively conspicuous and enduring regularities. So it is with the study of language use. The current limits of those regularities fix the current limits of the cleared area called 'meaning'. So where 'the explanatory power of standard sense' comes to an end, so does semantics. (1987, 286)

Rorty's analogy captures the difference between the Hessian approach to meaning that we have defended and a more traditional view that seeks to idealise away from minor perturbations in meaning change. The exciting possibility that large language models introduce is that such idealizations may no longer be necessary.

**Acknowledgements**: Thanks to audiences at the Digital Pragmatics and Epistemology Reading Group 2023, the Experimental Philosophy Group at the University of East Anglia 2024, the Digital Humanities and AI conference at the University of Reading 2024, and the Dartmouth College Neukom Workshop on Artificial Intelligence, Language, and Cognition. Thanks also to Kamil Lemanek for generous comments. Nat Hansen acknowledges generous support from The Alexander von Humboldt Foundation and from a Research Fellowship from the University of Reading.